# Uncertainty Quantification for Scientific Machine Learning using Sparse Variational Gaussian Process Kolmogorov-Arnold Networks (SVGP KAN)

*Y. Sungtaek Ju*
*Department of Mechanical and Aerospace Engineering*
*UCLA*
*sungtaek.ju@ucla.edu*

## Abstract

Kolmogorov-Arnold Networks have emerged as interpretable alternatives to traditional multi-layer perceptrons. However, standard implementations lack principled uncertainty quantification capabilities essential for many scientific applications. We present a framework integrating sparse variational Gaussian process inference with the Kolmogorov-Arnold topology, enabling scalable Bayesian inference with computational complexity quasi-linear in sample size. Through analytic moment matching, we propagate uncertainty through deep additive structures while maintaining interpretability. We use three example studies to demonstrate the framework's ability to distinguish aleatoric from epistemic uncertainty: calibration of heteroscedastic measurement noise in fluid flow reconstruction, quantification of prediction confidence degradation in multi-step forecasting of advection-diffusion dynamics, and out-of-distribution detection in convolutional autoencoders. These results suggest Sparse Variational Gaussian Process Kolmogorov-Arnold Networks (SVGP-KANs) is a promising architecture for uncertainty-aware learning in scientific machine learning.

## 1. Introduction

The quantification of uncertainty in neural network predictions is an important requirement in scientific machine learning, medical diagnosis, autonomous systems, and engineering design. Traditional neural architectures based on multi-layer perceptrons provide point estimates without principled measures of confidence, limiting their deployment in applications where understanding model uncertainty is as important as prediction accuracy itself. While Bayesian neural networks offer a theoretical framework for uncertainty quantification, their computational demands and challenges in posterior inference have motivated the development of alternative approaches combining expressivity with tractable probabilistic reasoning.

The Kolmogorov-Arnold network [1–3] represents a major shift in neural architecture design by leveraging the Kolmogorov-Arnold representation theorem. This theorem establishes that any multivariate continuous function defined on a finite domain can be represented as a finite composition of continuous univariate functions and addition operations. Unlike conventional multi-layer perceptrons that employ fixed nonlinear activations at nodes, Kolmogorov-Arnold networks place learnable univariate functions on edges, creating an additive structure that enhances interpretability. The original implementation employs B-spline parameterizations for these univariate functions, enabling direct visualization and symbolic regression capabilities that have proven valuable in scientific discovery tasks.

Recent extensions have explored various formulations of Kolmogorov-Arnold networks across diverse application domains [3]. As an example, Bodner et al. [4] developed convolutional variants for computer vision tasks, demonstrating that the edge-based learning paradigm can be successfully utilized in such tasks. Zhang and Zhang [5] introduced graph-theory extensions for relational data, while Genet and Inzirillo [6] and Yang and Wang [7] proposed temporal adaptations for sequence modeling. Alternative basis functions also have been investigated. Xu et al. [8] replaced B-splines with Fourier bases for improved computational efficiency, Yang and Wang [7] explored rational functions to reduce parameter counts, Bozorgasl and Chen [9] employed wavelet decompositions for multi-resolution analysis, and Li [10] demonstrated equivalences between certain Kolmogorov-Arnold formulations and radial basis function networks. Zheng et al. [11] analyzed the theoretical properties of spline knots in B-spline-based implementations, deriving tight upper bounds on representational capacity and proposing free-knot adaptations to enhance expressive power while maintaining training stability.

Despite these advances, the deterministic nature of standard Kolmogorov-Arnold networks limits their applicability in uncertainty-critical domains. Hassan et al. (2024) attempted to address this limitation through Bayesian Kolmogorov-Arnold networks, placing probability distributions over B-spline coefficients. While this captures weight-space uncertainty, it does not fully capture the function-space uncertainty inherent in non-parametric Bayesian modeling. A previous study [12] proposed Gaussian process Kolmogorov-Arnold networks, directly modeling edge functions as drawn from Gaussian process priors. Their formulation provides principled function-space uncertainty but relies on exact inference, requiring inversion of $N \times N$ covariance matrices with $O(N^3)$ computational complexity and thereby limiting scalability. Zou and Yan [13] recently proposed probabilistic formulations using sparsified deep Gaussian processes with additive kernels, though their focus remained on theoretical properties rather than practical applications requiring fine-grained spatial uncertainty quantification.

We introduce a framework combining sparse variational inference with Kolmogorov-Arnold network topology to achieve quasi-linear scalability while preserving interpretability and uncertainty quantification capabilities. Our approach employs inducing point approximations following [14] and [15], reducing per-sample complexity from $O(N^3)$ to $O(NM^2)$ where $M \ll N$ represents the number of inducing points. The univariate structure of Kolmogorov-Arnold edges enables independent variational approximations across edge functions, avoiding the computational challenges associated with high-dimensional variational inference in general deep Gaussian processes. Through analytic moment matching [16], we propagate input uncertainty through layers via closed-form expressions, maintaining computational tractability while quantifying both aleatoric uncertainty arising from measurement noise and epistemic uncertainty reflecting model limitations.

The remainder of the manuscript is organized as follows. First, we discuss a mathematical framework for sparse variational Gaussian process inference adapted to Kolmogorov-Arnold topology, presenting tractable evidence lower bounds and gradient expressions for stochastic optimization. Next, we present application of the framework to study three different example problems with relevance to fluid mechanics, scalar transport, and anomaly detection in biomedicine and other disciplines. In the first example study, we demonstrate calibration of spatially varying aleatoric uncertainty in a fluid dynamics problem with heteroscedastic measurement noise. In the second example study, we quantify epistemic uncertainty growth in multi-step forecasting of advection-diffusion transport, showing that prediction confidence naturally degrades with forecast horizon due to the accumulation of model approximation errors in sequential predictions. In the third example study, we establish out-of-distribution detection capabilities through integration with convolutional autoencoders.

## 2. Mathematical Framework

### 2.1 Kolmogorov-Arnold Networks and Probabilistic Edge Functions

A Kolmogorov-Arnold network (KAN) transforms an input vector $\mathbf{x} \in \mathbb{R}^{D_{\text{in}}}$ to an output $\mathbf{y} \in \mathbb{R}^{D_{\text{out}}}$ through a composition of layers, where each layer implements an additive aggregation of univariate transformations. For a single layer mapping dimension $D_{\text{in}}$ to dimension $D_{\text{out}}$, the $j$-th output component is computed as

$$y_j = \sum_{i=1}^{D_{\text{in}}} \phi_{j,i}(x_i) + b_j$$

where $\phi_{j,i}: \mathbb{R} \to \mathbb{R}$ denotes a univariate function associated with the edge connecting input dimension $i$ to output dimension $j$, and $b_j$ represents an optional bias term. This formulation differs fundamentally from conventional multi-layer perceptrons, which apply fixed activation functions after affine transformations. The Kolmogorov-Arnold structure places learnable nonlinearities on edges rather than nodes, enabling direct interpretation of individual feature contributions through visualization of univariate functions.

In deterministic implementations, $\phi_{j,i}$ are parameterized using B-splines, Chebyshev polynomials, Fourier bases, or rational functions. We instead model each edge function as a Gaussian process, introducing a probabilistic interpretation that naturally quantifies uncertainty. Specifically, we place a zero-mean Gaussian process prior over each $\phi_{j,i}$:

$$\phi_{j,i} \sim \mathcal{GP}(0, k(\cdot,\cdot; \theta_{j,i}))$$

where $k: \mathbb{R} \times \mathbb{R} \to \mathbb{R}$ denotes a kernel function parameterized by hyperparameters $\theta_{j,i}$ that may be learned during training. This construction transforms the deterministic Kolmogorov-Arnold layer into a probabilistic mapping where the output $y_j$ becomes a random variable with uncertainty quantified through its posterior distribution.

The choice of kernel function determines the inductive bias of the model. For universal approximation with smooth functions, we primarily employ the squared exponential (radial basis function) kernel:

$$k_{\text{SE}}(x, x'; \sigma_f^2, \ell) = \sigma_f^2 \exp\left(-\frac{(x-x')^2}{2\ell^2}\right)$$

where $\sigma_f^2$ controls the prior variance of function values and $\ell$ represents the lengthscale determining the correlation structure. Alternative kernels such as the Matérn family may be employed when modeling phenomena with specific smoothness properties or when incorporating prior knowledge about function regularity. However, the closed-form KL divergence calculations presented in Section 2.2 are specific to the RBF kernel; extensions to other kernel families require deriving the corresponding KL expressions.

## 2.2 Sparse Variational Inference for Scalable Training

Exact Gaussian process inference requires computing and inverting the $N \times N$ Gram matrix $\mathbf{K}$ whose elements are $K_{nm} = k(x_n, x_m)$ for training inputs $\{x_n\}_{n=1}^{N}$. The computational complexity of $O(N^3)$ for matrix inversion and $O(N^2)$ storage requirements prohibit application to large datasets. We employ sparse variational approximations to reduce this complexity while maintaining the theoretical benefits of Bayesian inference.

For each edge function $\phi_{j,i}$, we introduce a set of $M$ inducing inputs $\mathbf{Z}_{j,i} = \{z_m\}_{m=1}^{M}$ where typically $M \ll N$. The inducing inputs represent pseudo-datapoints in the input space, and their corresponding function values $\mathbf{u}_{j,i} = \phi_{j,i}(\mathbf{Z}_{j,i})$ are termed inducing variables. Following [14], the key insight is that conditioning on $\mathbf{u}_{j,i}$ renders function values at different data points approximately independent:

$$p(\phi_{j,i}(\mathbf{X}) \mid \mathbf{u}_{j,i}) \approx \prod_{n=1}^{N} p\left(\phi_{j,i}(x_n) \mid \mathbf{u}_{j,i}\right)$$

This factorization reduces the computational burden from $O(N^3)$ to $O(NM^2)$, enabling application to large datasets.

We employ the variational framework [15] to approximate the true posterior $p(\mathbf{u}_{j,i} \mid \mathbf{y})$ with a parametric variational distribution $q(\mathbf{u}_{j,i}) = \mathcal{N}(\mathbf{m}_{j,i}, \mathbf{S}_{j,i})$ where $\mathbf{m}_{j,i} \in \mathbb{R}^M$ and $\mathbf{S}_{j,i} \in$ $\mathbb{R}^{M \times M}$ are variational parameters to be optimized. The variational posterior for the function value at an arbitrary input $x$ is obtained through the conditioning equations:

$$q(\phi_{j,i}(x)) = \int p(\phi_{j,i}(x) \mid \mathbf{u}_{j,i}) q(\mathbf{u}_{j,i}) d\mathbf{u}_{j,i}$$

yielding a Gaussian distribution with mean and variance:

$$\mu_{j,i}(x) = \mathbf{k}_x^\top \mathbf{K}_{ZZ}^{-1} \mathbf{m}_{j,i}$$
$$\sigma_{j,i}^2(x) = k(x,x) - \mathbf{k}_x^\top \mathbf{K}_{ZZ}^{-1} (\mathbf{K}_{ZZ} - \mathbf{S}_{j,i}) \mathbf{K}_{ZZ}^{-1} \mathbf{k}_x$$

where $\mathbf{k}_x = [k(x,z_1), \ldots, k(x,z_M)]^\top$ and $\mathbf{K}_{ZZ}$ denotes the $M \times M$ kernel matrix evaluated at inducing inputs.

The evidence lower bound (ELBO) provides a tractable objective for stochastic optimization. For a single edge function with scalar outputs, the ELBO decomposes as:

$$\mathcal{L}_{\text{ELBO}} = \mathbb{E}_{q(\phi)}[\log p(\mathbf{y} \mid \phi)] - \lambda \cdot \text{KL}[q(\mathbf{u}) \parallel p(\mathbf{u})]$$

where $\lambda$ is a hyperparameter controlling the relative weighting of the KL divergence term. The KL divergence term admits a closed form for Gaussian distributions:

$$\text{KL}[q(\mathbf{u}) \parallel p(\mathbf{u})] = \frac{1}{2}\left(\text{tr}(\mathbf{K}_{ZZ}^{-1} \mathbf{S}) + \mathbf{m}^\top \mathbf{K}_{ZZ}^{-1} \mathbf{m} - M + \log \frac{|\mathbf{K}_{ZZ}|}{|\mathbf{S}|}\right)$$

The KL weight $\lambda$ represents a critical design choice that balances data fidelity against Bayesian regularization. Recent work on variational inference has established that $\lambda$ significantly impacts both predictive accuracy and uncertainty calibration [17,18]. Empirically, values in the range $\lambda \in [0.001, 0.1]$ have proven effective across diverse applications [19,20], with smaller values prioritizing data fit and larger values encouraging conservative uncertainty estimates. In our experiments, we employ $\lambda$ of the order of = 0.01 to provide a balanced trade-off between fitting capacity and regularization.

The expected log-likelihood term must be approximated numerically but can be efficiently estimated using mini-batches, enabling stochastic gradient descent optimization. The mini-batch approximation ensures that per-iteration complexity scales as $O(BM^2N)$ where $B$ denotes batch size, achieving linear scaling in the full dataset size $N$.

## 2.3 Predictive Variance Decomposition and Out-of-Distribution Calibration

An important consideration in sparse variational Gaussian process inference is the proper treatment of predictive variance, particularly for out-of-distribution (OOD) detection. The exact predictive variance at a test point $x_*$ can be decomposed into two geometrically interpretable components [14,21]:

$$\text{Var}(f_*) = \underbrace{\mathbf{k}_{x_*}^\top \mathbf{K}_{ZZ}^{-1} \text{Cov}(\mathbf{u}) \mathbf{K}_{ZZ}^{-1} \mathbf{k}_{x_*}}_{\text{Projected Variance } (V_{\text{proj}})} + \underbrace{\left[k(x_*, x_*) - \mathbf{k}_{x_*}^\top \mathbf{K}_{ZZ}^{-1} \mathbf{k}_{x_*}\right]}_{\text{Orthogonal Variance } (V_{\text{orth}})}$$

The projected variance $V_{\text{proj}}$ quantifies uncertainty transmitted through the inducing point subspace. That is, it captures how posterior uncertainty over $\mathbf{u}$ propagates to predictions at $x_*$. The orthogonal variance $V_{\text{orth}}$, identified [21] as the Nyström approximation error, measures information lost by the sparse representation: the component of $f_*$ that lies outside the span of the inducing functions.

Standard implementations of moment-matching in deep GP architectures often retain only $V_{\text{proj}}$, implicitly assuming the inducing points provide sufficient coverage of the input domain. This omission leads to variance collapse for out-of-distribution inputs: as a test point $x_*$ moves far from all inducing locations $\mathbf{Z}$, the cross-covariance $k(x_*, \mathbf{Z}) \to \mathbf{0}$, causing projected variance to vanish regardless of true epistemic uncertainty. We address this by incorporating the Nyström approximation error $V_{\text{orth}} = k(x_*, x_*) - k(x_*, \mathbf{Z})K_{\mathbf{ZZ}}^{-1}k(\mathbf{Z}, x_*)$ into our predictive variance, approximated efficiently via the $\psi$-statistics as $V_{\text{orth}} \approx \sigma_f^2(1 - \sum_m \psi_1^2/\sigma_f^2)$. This guarantees that as $x_* \to \infty$, predictive variance recovers the prior $\sigma_f^2$, providing a geometric safeguard against silent extrapolation failures essential for the out-of-distribution detection demonstrated in Study C.

## 2.4 Analytic Moment Matching for Uncertainty Propagation

A fundamental challenge in deep Gaussian process architectures concerns the propagation of uncertainty through layers. When the input to a Gaussian process is itself a random variable with distribution $p(x)$, evaluating the posterior predictive distribution requires computing expectations over this input distribution. For arbitrary input distributions and kernels, these expectations are generally intractable. However, the univariate structure of Kolmogorov-Arnold edges and the additive aggregation operation enable tractable approximations through analytic moment matching.

Consider a Gaussian process $\phi$ with squared exponential kernel parameterized by signal variance $\sigma_f^2$ and lengthscale $\ell$. Suppose the input $x$ follows a Gaussian distribution $x \sim \mathcal{N}(\mu_x, \sigma_x^2)$. We seek to compute the mean and variance of the output distribution $p(y) = \int p(y \mid x)p(x)dx$ where $p(y \mid x)$ represents the posterior predictive distribution of the Gaussian process.

Following [16], the expected posterior mean can be computed as:

$$\mathbb{E}_{p(x)}[\mu_\phi(x)] = \mathbf{m}^\top \mathbf{K}_{\mathbf{ZZ}}^{-1} \boldsymbol{\psi}_1$$

where the vector $\boldsymbol{\psi}_1 \in \mathbb{R}^M$ has components:

$$[\boldsymbol{\psi}_1]_m = \mathbb{E}_{p(x)}[k(x, z_m)] = \sigma_f^2 \left(\frac{\ell^2}{\ell^2 + \sigma_x^2}\right)^{1/2} \exp\left(-\frac{(z_m - \mu_x)^2}{2(\ell^2 + \sigma_x^2)}\right)$$

This expression reveals a regularization property inherent in uncertainty propagation. As input uncertainty $\sigma_x^2$ increases, the effective lengthscale becomes $\sqrt{\ell^2 + \sigma_x^2}$, resulting in smoother effective functions. This automatic smoothing prevents overconfident extrapolation by increasing prediction uncertainty in regions where inputs are uncertain.

The variance of the output distribution comprises two terms reflecting distinct sources of uncertainty:

$$\mathbb{V}_{p(x)}[y] = \mathbb{E}_{p(x)}[\sigma_\phi^2(x)] + \mathbb{V}_{p(x)}[\mu_\phi(x)]$$

The first term captures epistemic uncertainty arising from finite training data, while the second reflects uncertainty propagation from stochastic inputs. For the additive structure of Kolmogorov-Arnold layers where $y = \sum_{i=1}^{D_{\text{in}}} \phi_i(x_i)$, the total output variance under a mean-field (layer-wise independence) approximation becomes:

$$\mathbb{V}[y] = \sum_{i=1}^{D_{\text{in}}} \left( \mathbb{E}[\sigma_{\phi_i}^2(x_i)] + \mathbb{V}[\mu_{\phi_i}(x_i)] \right)$$

This additive decomposition enables efficient propagation through deep networks while maintaining interpretability, as uncertainty contributions from individual input dimensions can be isolated and analyzed.

As discussed in Section 2.3, the epistemic uncertainty term $\mathbb{E}[\sigma_\phi^2(x)]$ incorporates both projected and orthogonal variance components, ensuring that predictive uncertainty correctly increases for inputs far from the inducing point support

## 2.5 Observation Noise and Aleatoric Uncertainty

In many scientific applications, measurements are corrupted by noise whose intensity varies spatially or temporally. Heteroscedastic observation noise, where variance depends on input location or system state, requires explicit modeling to avoid biased predictions and miscalibrated uncertainties. We extend the framework to incorporate input-dependent observation noise through a secondary Gaussian process.

Let $f(\mathbf{x})$ denote the latent function of interest and $\sigma_{\text{noise}}^2(\mathbf{x})$ represent the observation noise variance at location $\mathbf{x}$. The likelihood becomes:

$$p(y \mid \mathbf{x}, f, \sigma_{\text{noise}}^2) = \mathcal{N}(y; f(\mathbf{x}), \sigma_{\text{noise}}^2(\mathbf{x}))$$

To ensure positivity, we model the log-variance through a separate Gaussian process:

$$\log \sigma_{\text{noise}}^2(\mathbf{x}) \sim \mathcal{GP}(m_{\text{noise}}(\mathbf{x}), k_{\text{noise}}(\mathbf{x}, \mathbf{x}'))$$

where $m_{\text{noise}}$ represents a mean function (often set to a learnable constant) and $k_{\text{noise}}$ denotes a kernel function that may differ from the kernel used for the latent function.

During training, both the latent function and noise variance are inferred jointly through the variational framework. The expected log-likelihood term in the ELBO becomes:

$$\mathbb{E}_{q(f)q(\log \sigma^2)} \left[ \sum_{n=1}^{N} \log \mathcal{N}(y_n; f(\mathbf{x}_n), \sigma^2(\mathbf{x}_n)) \right]$$

This expectation can be approximated using sampling or quadrature. The resulting framework naturally separates aleatoric uncertainty (inherent stochasticity in observations) from epistemic uncertainty (model uncertainty due to finite data), providing calibrated confidence intervals that correctly expand in low-data regions and contract where observations are plentiful and precise.

## 2.6 Computational Complexity and Scalability

The computational advantages of Sparse Variational Gaussian Process Kolmogorov-Arnold Networks (SVGP KANs) arise from the factorization of edge functions and inducing point approximations. Consider a network with $L$ layers, where layer $l$ maps dimension $D_l$ to $D_{l+1}$. The total number of edge functions across the network is $\sum_{l=1}^{L} D_l D_{l+1}$. For each edge function, maintaining variational parameters requires storing $M$ inducing inputs, an $M$-dimensional mean vector, and an $M \times M$ covariance matrix, totaling $O(M + M^2)$ parameters per edge.

Forward propagation through a single layer requires computing posterior means and variances for $D_{l+1}$ outputs, each involving $D_l$ edge evaluations. Each edge evaluation requires matrix-vector products with $O(M^2)$ complexity, yielding $O(D_l D_{l+1} M^2)$ per-layer cost. For typical architectures where $M$ ranges from 20 to 100, this remains tractable even for networks with thousands of parameters. Importantly, the cost scales linearly with training set size $N$ when using mini-batch stochastic optimization, contrasting with $O(N^3)$ scaling in exact Gaussian process inference.

The inducing inputs $\mathbf{Z}_{j,i}$ may be initialized uniformly across the input domain or through k-means clustering of training data. During optimization, these inducing inputs can be treated as learnable parameters, enabling the model to automatically position pseudo-datapoints in regions of high variability or importance. Gradient-based optimization of inducing locations has been shown to improve approximation quality in sparse Gaussian process literature.

## 3. Experimental Design and Implementation

### 3.1 Architecture Specifications

We implement the sparse variational Gaussian process Kolmogorov-Arnold network framework in PyTorch. The core architectural component, termed GPKANLayer, replaces standard linear transformations with sparse variational Gaussian process mappings. Each layer maintains inducing inputs stored as learnable parameters of dimension $(D_{\text{in}} \times D_{\text{out}}) \times M$, variational means of the same dimension, and lower-triangular Cholesky factors parameterizing variational covariances.

For fluid dynamics applications requiring convolutional architectures, we construct encoder-decoder networks where dense Kolmogorov-Arnold layers form bottlenecks between convolutional feature extractors and generators. The encoder processes input fields through convolutional layers with kernel sizes adapted to the spatial scale of features, followed by flattening and projection through a GPKANLayer bottleneck. The decoder performs the inverse operation, mapping latent representations back to the spatial domain through transposed convolutions.

Kernel functions are implemented with learnable hyperparameters including lengthscales and signal variances. We employ log-parameterization for positive hyperparameters to ensure valid parameter spaces during unconstrained optimization. Initial lengthscale values are set based on the typical range of input features, with signal variances initialized to unity. For heteroscedastic noise modeling, we introduce separate Gaussian process priors over log-noise-variance, enabling the model to learn spatially varying observation uncertainty from data.

### 3.2 Study A: Calibration of Heteroscedastic Measurement Uncertainty

The first study evaluates the framework's ability to identify and calibrate spatially varying observation noise in fluid velocity field reconstruction. We construct a synthetic dataset simulating noisy measurements of a two-dimensional incompressible flow field, where measurement uncertainty may vary spatially according to sensor quality or environmental conditions.

The ground truth velocity field is generated as a divergence-free vector field satisfying the incompressibility constraint $\nabla \cdot \mathbf{v} = 0$. We employ a stream function formulation where the velocity components are derived as:

$$v_x(x,y) = \frac{\partial \psi}{\partial y}, \quad v_y(x,y) = -\frac{\partial \psi}{\partial x}$$

The stream function $\psi$ is constructed as a superposition of smooth basis functions:

$$\psi(x,y) = \sum_{k=1}^{K} A_k \sin(n_k \pi x / L)\sin(m_k \pi y / L)$$

where $A_k$ denote amplitudes, $(n_k, m_k)$ represent wavenumbers, and $L$ sets the domain size. This formulation guarantees smooth velocity fields with controllable complexity.

Observation noise is generated according to a spatially varying variance function designed to simulate sensor networks where measurement precision degrades in certain regions. We define the noise variance as:

$$\sigma_{\text{true}}^2(x,y) = \sigma_{\text{base}}^2 \left(1 + A_{\text{noise}} \exp\left(-\frac{(x-x_c)^2 + (y-y_c)^2}{2r^2}\right)\right)$$

This creates localized high-noise regions centered at $(x_c, y_c)$ with characteristic radius $r$. Training data consists of velocity measurements corrupted by Gaussian noise with variance $\sigma_{\text{true}}^2$ at each location, while the model must simultaneously reconstruct the velocity field and infer the noise distribution from observations alone.

The network architecture comprises a three-layer design mapping two-dimensional spatial coordinates $(x, y)$ to two-dimensional velocity predictions $(\hat{v}_x, \hat{v}_y)$. Each hidden layer employs 5 neurons with 20 inducing points per edge function. The observation noise is modeled through a parallel Gaussian process predicting log-noise-variance, using squared exponential kernels with separate lengthscale hyperparameters to capture potentially different spatial correlation structures between the latent velocity field and noise distribution.

Training proceeds through mini-batch stochastic gradient descent over 1000 epochs with batch size 16, using on-the-fly data generation on a 64x64 spatial grid. The loss function combines the negative log-likelihood under the learned heteroscedastic noise model with KL divergence regularization weighted by $\lambda = 0.01$. This KL weight provides balanced regularization, ensuring well-calibrated uncertainty estimates without over-conservative predictions. Evaluation is performed on 100 independent test samples, totaling 409,600 prediction points.

### 3.3 Study B: Epistemic Uncertainty in Multi-Step Forecasting

The second study examines epistemic uncertainty growth in sequential predictions of advection-diffusion transport phenomena. Unlike the first study where uncertainty primarily reflects aleatoric measurement noise, this investigation focuses on model uncertainty arising from approximation errors in the neural network that compound over multiple prediction steps, even when the underlying physical dynamics remain stable.

We simulate two-dimensional advection-diffusion dynamics governing the evolution of a scalar temperature field $T(\mathbf{x}, t)$ under the partial differential equation:

$$\frac{\partial T}{\partial t} = -\mathbf{u} \cdot \nabla T + \kappa \nabla^2 T + S(\mathbf{x}, t)$$

where $\mathbf{u}(\mathbf{x}, t)$ denotes a prescribed velocity field, $\kappa$ represents the thermal diffusivity, and $S$ incorporates source and sink terms. The velocity field is assumed to evolve slowly in time according to a time-dependent stream function:

$$\psi(\mathbf{x}, t) = \sin(2x + \omega t)\cos(y) + 0.5\cos(x)\sin(2y - 0.5\omega t)$$

where $\omega = 0.1$ controls the temporal evolution rate. This creates a gradually varying flow pattern where different initial conditions experience different advective transport histories.

We again use a 64x64 spatial grid with periodic boundary conditions. Forward Euler time integration with timestep $\Delta t = 0.1$ advances the system for 15 steps, providing ground truth trajectories. To create an advection-dominated regime that tests the model's predictive capabilities, we set the advection velocity magnitude to 0.5 and the diffusivity to $\kappa = 0.002$, yielding a Péclet number $Pe \approx 1571$. The source term amplitude is 0.02. The model is trained solely on single-step transitions, that is, given temperature field $T_n$, we predict $T_{n+1}$. This training protocol reflects realistic scenarios where only short-term measurements are available, yet long-term forecasts are required.

The physical system's sensitivity to initial conditions was tested independently by evolving two trajectories with 0.1% initial perturbation using the ground truth PDE solver. The mean difference at $t = 14$ remained essentially constant at 1.0x the initial difference, confirming that the system is stable rather than chaotic. This stability analysis is critical for interpreting subsequent results: any observed uncertainty growth in model predictions reflects epistemic limitations of the neural network approximation, not sensitive dependence on initial conditions in the underlying physics.

The SVGP-KAN architecture employs a convolutional encoder-decoder structure. The encoder uses two convolutional layers with ReLU activations, reducing spatial dimensions. A GPKANLayer bottleneck with 24 base channels and 20 inducing points per edge compresses the representation. The decoder mirrors the encoder structure using transposed convolutions to reconstruct the temperature field. Training proceeds over 1000 epochs using mean squared error loss augmented with KL divergence regularization weighted by $\lambda = 0.01$, employing the Adam optimizer with learning rate $2 \times 10^{-3}$.

Uncertainty quantification in multi-step forecasting is performed through ensemble rollout. We generate an ensemble of $n_{\text{ens}} = 10$ initial conditions by perturbing a reference initial state with 1% Gaussian noise. Each ensemble member is propagated forward for 15 steps using the posterior mean prediction (deterministic propagation) at each timestep, with predictions clipped to the physical domain $[-2,2]$ to prevent numerical instabilities. The ensemble spread, measured through pixelwise standard deviation across ensemble members, quantifies prediction uncertainty at each temporal horizon.

### 3.4 Study C: Out-of-Distribution Detection in Convolutional Autoencoders

The third study evaluates anomaly detection capabilities by integrating SVGP-KANs as bottleneck layers in convolutional autoencoders. The task requires distinguishing in-distribution inputs that the model was trained on from anomalous out-of-distribution inputs that should trigger high uncertainty alerts.

This example study can be extended and applied to a wide variety of applications in science and technology, including bio imaging. As a notional demonstration, we use the readily accessible MNIST dataset of handwritten digits, training the autoencoder exclusively on images of the digit "0". The architecture follows a standard encoder-decoder design: the encoder applies two convolutional layers (8 filters, then 16 filters, each with 3x3 kernels and stride 2) followed by ReLU activations, reducing spatial dimensions from 28 x 28 to 7 x 7. The flattened feature map of dimension 16 x 7 x 7 = 784 is projected through a GPKANLayer bottleneck to a 6-dimensional latent code. The decoder mirrors the encoder, first expanding the latent code to 784 dimensions through a linear layer, reshaping to spatial dimensions, then applying transposed convolutions to reconstruct 28x28 images. The GPKANLayer bottleneck employs 20 inducing points per edge with RBF kernels. Lengthscale and signal variance hyperparameters are initialized to moderate values and remain trainable, allowing the model to learn appropriate uncertainty scales from data.

Training proceeds over 70 epochs using mean squared error reconstruction loss augmented with KL divergence regularization weighted by $\lambda = 0.01$. The orthogonal variance correction described in Section 2.3 ensures that predictive variance correctly reverts to the prior when the encoder projects out-of-distribution features into regions far from inducing points. This enables principled Bayesian inference while maintaining out-of-distribution detection capabilities

At test time, we present both normal "0" digit images and anomalous "7" digit images. For each input, we compute the reconstruction error and the bottleneck uncertainty, defined as the sum of predicted variances across the 6 latent dimensions. The hypothesis is that anomalous inputs will yield both higher reconstruction errors (the decoder cannot accurately generate "7" patterns from latent codes trained on "0" patterns) and higher bottleneck uncertainties. The encoder encounters feature combinations unseen during training, triggering high epistemic uncertainty in the Gaussian process edges due to reversion to the prior variance through the orthogonal variance mechanism.

Quantitative evaluation measures the anomaly score ratio, that is, the uncertainty on anomalous inputs divided by the uncertainty on normal inputs. A well-calibrated model should produce ratios significantly greater than unity, indicating clear discrimination. We further evaluate discrimination quality through receiver operating characteristic (ROC) analysis, computing the area under the curve (AUC) to assess overall separation quality. We also examine reconstruction quality visually, testing whether anomalous inputs produce characteristic degraded or smoothed reconstructions reflecting the model's attempt to project out-of-distribution features onto the learned manifold of normal data.

## 3.5 Computational Overhead

The computational overhead introduced by sparse variational inference remains modest relative to deterministic implementations. For edge functions with $M = 20$ inducing points, the variational parameters (mean vector and covariance matrix) add approximately $M + M^2 = 420$ parameters per edge. A deterministic B-spline implementation with grid size $G = 20$ and order $K = 3$ stores $(G + K)$ coefficients per edge, totaling 23 parameters. Thus, the probabilistic formulation increases parameter count by roughly 18-fold per edge. However, this comparison understates the computational advantage: deterministic B-splines require recursive evaluation during forward passes, while inducing point formulations employ simple matrix-vector products amenable to GPU parallelization. In practice, we observe training times within comparable ranges between deterministic and probabilistic implementations when using equivalent network architectures.

# 4. Results

## 4.1 Heteroscedastic Noise Calibration in Fluid Flow Reconstruction

The model's ability to infer spatially varying noise patterns was evaluated on synthetic fluid fields with heteroscedastic measurement uncertainty to demonstrate that KL-regularized training produces well-calibrated uncertainty estimates. The Pearson correlation between predicted uncertainty and actual absolute error was $\rho$=0.55, indicating statistically significant learning of heteroscedastic patterns. Note that this moderate correlation reflects the stochastic nature of the problem, that is, the predicted uncertainty represents expected error magnitude, not deterministic error values. The correlation is strong enough to demonstrate that the model has successfully learned to identify high-noise regions, yet not perfect as would be inappropriate for a system with inherent measurement noise.

Coverage statistics demonstrated excellent calibration across multiple confidence levels, less than 1 percentage points at ±1σ, ±2σ, and ±3σ. The calibration error at $2\sigma$ was 0.1 percentage points, well within acceptable bounds for practical uncertainty quantification. Standardized prediction errors $(y - \mu)/\sigma$ exhibited mean of -0.09 and standard deviation 1.01, closely approximating the ideal $\mathcal{N}(0,1)$ distribution expected for well-calibrated probabilistic predictions. The near-zero mean confirms unbiased predictions. Overall RMSE of 0.085 demonstrates accurate mean predictions alongside well-calibrated uncertainty estimates.

Figure 1 visualizes the results from the first study. The predicted uncertainty field exhibits spatial structure corresponding to the ground truth heteroscedastic noise pattern, with elevated uncertainty in the designated high-noise regions. The calibration scatter plot shows a strong linear relationship between predicted uncertainty and actual error. The deviation from perfect linearity reflects appropriate acknowledgment of irreducible aleatoric noise rather than miscalibration.

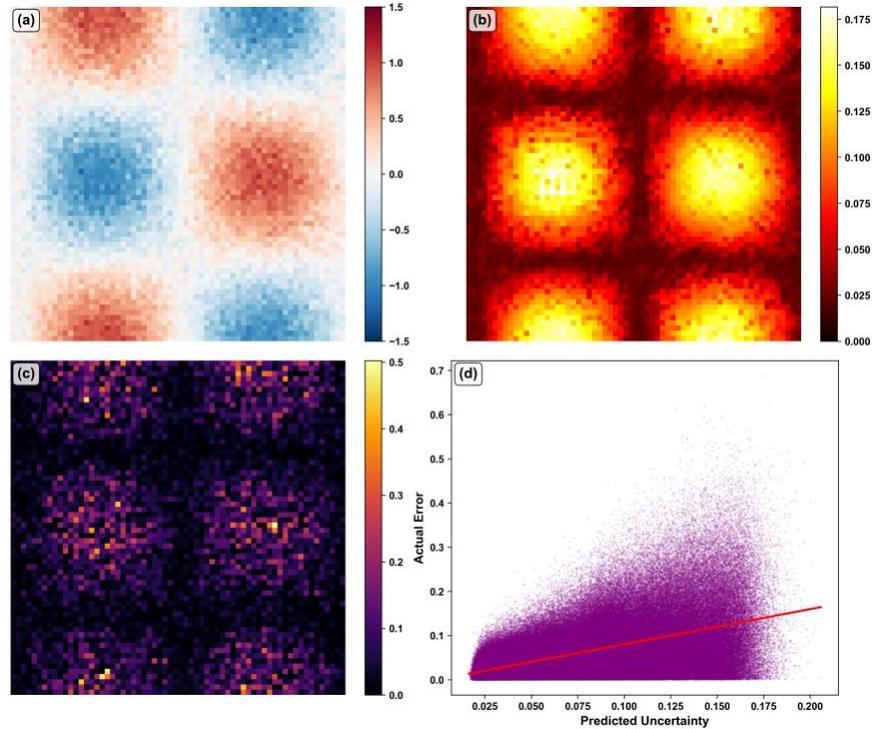

**Figure 1: Heteroscedastic Noise Calibration in Fluid Flow Reconstruction:** Spatially varying uncertainty calibration. (a) Prediction: Model reconstruction of velocity field showing spatial variation with characteristic high (red) and low (blue) regions. (b) Predicted Uncertainty: Spatially resolved uncertainty estimates $\sigma(x, y)$ (c) Absolute prediction errors $|y_{\text{true}} - y_{\text{pred}}|$. (d) Calibration Scatter Plot.

### 4.2 Multi-Step Prediction Uncertainty Growth in Advection-Diffusion

The time-dependent advection-diffusion system reveals characteristic epistemic uncertainty growth patterns in multi-step forecasting of stable dynamics. Figure 2 displays the evolution over a 15-step forecast horizon, showing ground truth concentration fields, ensemble mean predictions, and ensemble spread (epistemic uncertainty).

At the initial timestep $t = 0$, the ground truth exhibits localized hot and cold regions arranged in a regular pattern. The ensemble mean prediction closely matches this initial condition, as expected given the small 1% initial perturbations. The ensemble spread at $t = 0$ remains nearly uniform at low amplitude (black in visualization), reflecting only the initial perturbation magnitude.

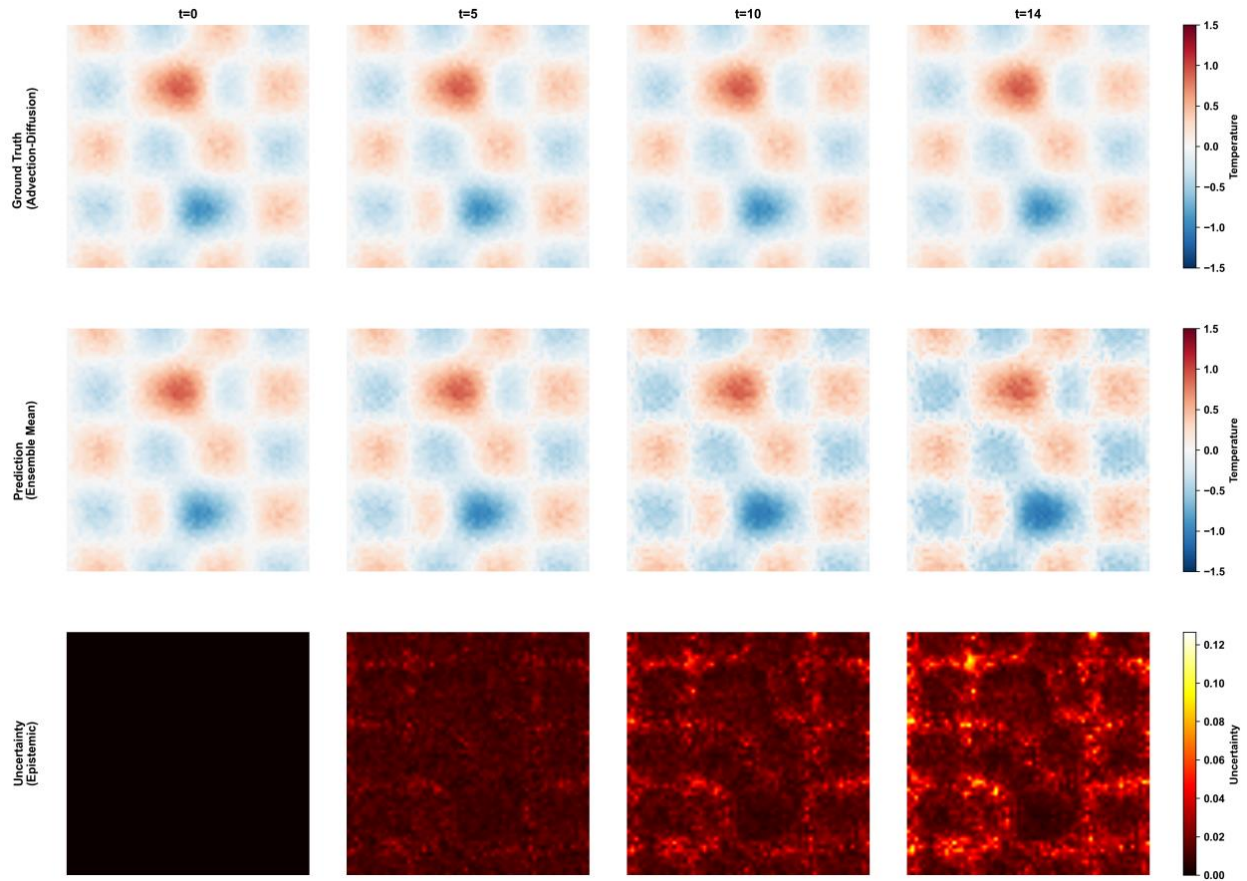

**Figure 2: Multi-Step Prediction Uncertainty Growth in Advection-Diffusion Dynamics:** Three-row visualization tracking epistemic uncertainty evolution over 15-step forecast horizon on 64x64 grid. Top row (Ground Truth), Center row (Prediction), and Bottom row (Uncertainty - Epistemic).

As the forecast progresses to $t = 5$, the ground truth concentration field evolves under combined diagonal advection and diffusion, with initial patterns transported and smoothed. The ensemble mean prediction tracks this evolution, though subtle differences begin to emerge. The ensemble spread increases, with localization beginning to appear in regions of steep gradients where small errors in predicted advection velocities compound.

At $t = 10$, divergence between predictions and ground truth becomes more pronounced. While large-scale structure remains qualitatively similar, quantitative differences in concentration magnitudes and spatial positions accumulate. The ensemble spread grows, exhibiting spatial heterogeneity with highest uncertainty concentrated along interfaces between hot and cold regions.

By the final forecast step $t = 14$, the ensemble mean prediction retains correlation with ground truth but exhibits noticeable smoothing and phase offsets. The uncertainty quantification accurately captures this degradation: ensemble spread reaches peak values with strong spatial organization indicating where model predictions have become unreliable (higher uncertainty).

Quantitative analysis reveals two distinct modes of uncertainty growth over the forecast horizon. The 2.4-fold temporal growth factor quantifies how model uncertainty accumulates over sequential predictions. Critically, as noted before, the physics sensitivity test performed on the ground truth solver confirmed that the underlying dynamics are stable. The observed 2.4x ensemble spread growth therefore reflects *model-induced epistemic uncertainty,* the neural network's imperfect approximation of the dynamics accumulates errors over multiple prediction steps, even though the underlying physics remains stable. This validates the framework's epistemic uncertainty quantification: the model correctly recognizes that its predictions become less reliable with increasing forecast horizon, assigning appropriately higher uncertainty to longer-term forecasts despite the stable underlying dynamics.

The 12.7x spatial variation ratio demonstrates that uncertainty is not uniform but concentrates in physically meaningful regions—specifically at flow interfaces where advective transport creates steep gradients that challenge the model's interpolation capabilities. This spatial structure provides actionable guidance for adaptive refinement strategies or targeted data collection in operational forecasting scenarios.

## 4.3 Anomaly Detection through Bottleneck Uncertainty

The convolutional autoencoder with SVGP-KAN bottleneck demonstrates robust out-of-distribution detection capabilities. Figure 3 shows visual comparison of normal and anomalous reconstructions.

The detection mechanism arises from the predictive variance decomposition described in Section 2.3. When presented with a "0" digit from the training distribution, the autoencoder reconstructs the input with high fidelity (top row, Fig. 3). The bottleneck uncertainty, measured as the sum of predicted variances across six latent dimensions, remains low for in-distribution samples. The variational posterior for inducing variables concentrates in regions of the GP input space corresponding to "0" features, and the projected variance $V_{\text{proj}}$ dominates the total predictive variance since inputs lie close to well-covered regions of the inducing point support.

In contrast, when processing a "7" digit unseen during training (bottom row, Fig. 3), reconstruction quality degrades substantially as the decoder attempts to force anomalous features onto the learned manifold of "0" patterns. Critically, bottleneck uncertainty increases substantially for out-of-distribution inputs because the encoder projects "7" features into regions far from any inducing points. In these regions, the cross-covariance $\mathbf{k}_{x_*}$ vanishes, causing the projected variance to approach zero. However, the orthogonal variance term $V_{\text{orth}} \to \sigma_f^2$ ensures that the total predictive variance reverts to the prior, providing the elevated uncertainty signal that enables anomaly detection. This geometric mechanism, guaranteed by the variance decomposition rather than dependent on specific hyperparameter choices, eliminates the need for explicit anomaly detection training while providing principled uncertainty estimates that identify unfamiliar inputs.

Evaluation across 50 test samples from each class confirms the discrimination capability of this uncertainty-based approach. With KL regularization weight $\lambda = 0.01$, we observe mean bottleneck uncertainty of $0.46 \pm 0.25$ for normal inputs compared to $2.49 \pm 1.31$ for anomalous inputs, yielding a mean uncertainty ratio of 5.4x. The ROC-AUC score of 0.964 indicates excellent discrimination quality, with 77.4% of anomalous samples exhibiting uncertainty exceeding the maximum observed on normal samples. The reconstruction MSE for anomalous inputs ($0.074 \pm 0.020$) is approximately 5.0x higher than for normal inputs ($0.015 \pm 0.006$), confirming degraded reconstruction quality on out-of-distribution data. These results demonstrate that the orthogonal variance correction enables robust OOD detection while maintaining principled Bayesian inference through non-zero KL regularization.

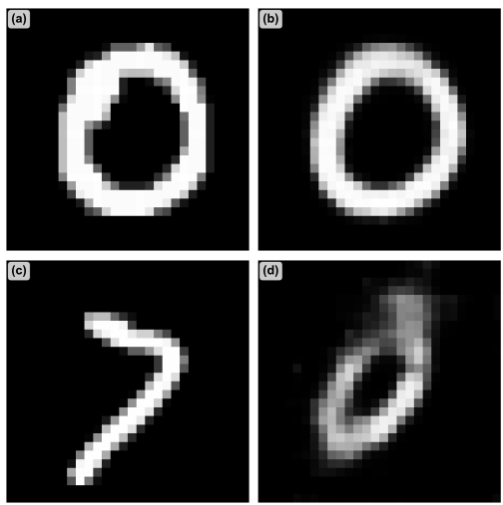

**Figure 3: Out-of-Distribution Detection via Bottleneck Uncertainty (Study C):** Comparison demonstrating anomaly detection through the orthogonal variance mechanism in SVGP-KAN bottleneck. Top row (Normal Input - Digit "0"), Bottom row (Anomalous Input - Digit "7"). Left: Input image. Right: Reconstruction.

## 5. Discussion

The three experimental studies demonstrate uncertainty quantification capabilities of our SVGP KAN across different application domains. First, we establish the framework's capacity for calibrating spatially varying aleatoric uncertainty in fluid flow reconstruction. The moderate correlation appropriately reflects irreducible aleatoric noise rather than deterministic prediction, validating that the model has learned heteroscedastic patterns without overfitting to individual noise realizations.

Second, we quantify epistemic uncertainty growth in multi-step forecasting of advection-diffusion dynamics, demonstrating that the model correctly identifies degrading prediction reliability over extended forecast horizons. The observed temporal uncertainty growth and large spatial variation occur despite stable underlying physics, confirming that the framework captures model-induced approximation errors rather than physical chaos. This distinction is critical: the neural network's imperfect dynamics approximation accumulates errors over sequential predictions, and the uncertainty estimates correctly reflect this degradation. The spatial concentration of uncertainty at flow interfaces provides actionable information for adaptive mesh refinement or targeted sensor deployment in operational settings.

Third, we establish out-of-distribution detection through bottleneck uncertainty in convolutional autoencoders. This capability emerges from the orthogonal variance term in the predictive variance decomposition, which guarantees that uncertainty reverts to the prior $\sigma_f^2$ when inputs lie far from the inducing point support. Unlike approaches that require omitting KL regularization to preserve OOD detection signals, the orthogonal variance mechanism provides a geometric guarantee that operates independently of the regularization weight $\lambda$. This enables a unified framework where all three studies employ principled Bayesian inference, achieving both well-calibrated in-distribution uncertainty (Studies A and B) and reliable out-of-distribution detection (Study C) without task-specific hyperparameter adjustments.

The computational scalability achieved through sparse variational approximations potentially enables application to large datasets while maintaining $O(NM^2)$ complexity with $M \ll N$. This represents a substantial improvement over exact Gaussian process inference ($O(N^3)$) without sacrificing the theoretical benefits of function-space uncertainty quantification. The additive structure of Kolmogorov-Arnold networks enables independent variational approximations across edges, avoiding the factorial complexity that would arise in general deep Gaussian process architectures.

Limitations and future directions also warrant discussion. First, the squared exponential kernel employed throughout assumes smooth functions; applications involving discontinuities or sharp transitions may benefit from Matérn kernels or piecewise-defined alternatives. Second, the inducing point count represents a trade-off between approximation quality and computational cost; automatic relevance determination schemes could adaptively allocate inducing points based on local function complexity. Third, the framework currently assumes Gaussian likelihoods; extensions to Poisson, Bernoulli, or heavy-tailed observation models would broaden applicability to count data, classification, and robust regression. Fourth, while we demonstrate uncertainty propagation through feedforward architectures, extensions to recurrent structures for temporal dynamics or graph neural networks for relational data are also promising directions. Finally, while the orthogonal variance correction ensures geometric consistency of uncertainty estimates independent of the KL weight, the choice of $\lambda$ still influences the balance between data fidelity and prior regularization; future work could investigate automatic adaptation schemes that dynamically adjust $\lambda$ during training based on validation set calibration metrics.

# 6. Conclusion

We have presented a framework for integrating sparse variational Gaussian process inference with Kolmogorov-Arnold network topology, enabling scalable Bayesian inference with $O(NM)$ training complexity while maintaining interpretability through additive edge-function structures. The predictive variance decomposition into projected and orthogonal components ensures both well-calibrated aleatoric uncertainty quantification for in-distribution predictions and principled epistemic uncertainty estimation that correctly reverts to the prior for out-of-distribution inputs.

The SVGP KAN architecture addresses the limitation of deterministic Kolmogorov-Arnold implementations (lack of principled uncertainty quantification) while maintaining flexibility across diverse application domains. It is promising as a practical framework for uncertainty-aware scientific computing.

Extensions to multi-output regression, non-Gaussian likelihoods, and temporal dynamics modeling will broaden the applicability of this framework across scientific disciplines requiring reliable uncertainty estimates under incomplete information, noisy observations, and distribution shift.

**Data Availability**

The code library and scripts used to generate the results are available in https://github.com/sungjuGit/svgp-kan.